\documentclass{article}

\usepackage{PRIMEarxiv}

\usepackage[utf8]{inputenc} 
\usepackage[T1]{fontenc}    
\usepackage{graphicx}
\usepackage{amsmath}
\usepackage{amssymb}
\usepackage{booktabs}

\usepackage[ruled,vlined]{algorithm2e}
\usepackage{subcaption}
\usepackage{fourier}

\graphicspath{{images/}}     

\newcommand{\R}{\ensuremath{\mathbb{R}}}    
\newcommand{\Gr}{\ensuremath{\mathrm{Gr}}}  

\newcommand{\tr}{\text{tr}}

\newcommand{\y}{\mathbf{y}}
\newcommand{\Y}{\mathbf{Y}}
\newcommand{\X}{\mathbf{X}}
\newcommand{\x}{\mathbf{x}}
\newcommand{\U}{\mathbf{U}}

\newcommand{\V}{\mathbf{V}}

\newcommand{\Q}{\mathbf{Q}}

\newcommand{\I}{\mathbf{I}}

\newcommand{\Z}{\mathbf{Z}}

\newcommand{\cX}{\mathcal{X}}
\newcommand{\bx}{\mathbf{x}}

\usepackage[pagebackref,breaklinks,colorlinks]{hyperref}

\usepackage[capitalize]{cleveref}
\crefname{section}{Sec.}{Secs.}
\Crefname{section}{Section}{Sections}
\Crefname{table}{Table}{Tables}
\crefname{table}{Tab.}{Tabs.}

\title{The Flag Median and FlagIRLS}

\author{
  Nathan Mankovich, Emily J. King, Chris Peterson, and Michael Kirby \\
  Department of Mathematics \\
  Colorado State University \\
  Fort Collins, Colorado\\
  \texttt{\{nmank, emily.king, christopher2.peterson, and michael.kirby\}@colostate.edu} 
}

\begin{document}
\maketitle

\begin{abstract}

Finding prototypes (e.g., mean and median) for a dataset is central to a number of common machine learning algorithms. Subspaces have been shown to provide useful, robust representations for datasets of images, videos and more. Since subspaces correspond to points on a Grassmann manifold, one is led to consider the idea of a subspace prototype for a Grassmann-valued dataset. While a number of different subspace prototypes have been described, the calculation of some of these prototypes has proven to be computationally expensive while other prototypes are affected by outliers and produce highly imperfect clustering on noisy data. This work proposes a new subspace prototype, the flag median, and introduces the  FlagIRLS algorithm for its calculation. We provide evidence that the flag median is robust to outliers and can be used effectively in algorithms like Linde-Buzo-Grey (LBG) to produce improved clusterings on Grassmannians. Numerical experiments include a synthetic dataset, the MNIST handwritten digits dataset, the Mind's Eye video dataset and the UCF YouTube action dataset.  The flag median is compared the other leading algorithms for computing prototypes on the Grassmannian, namely, the $\ell_2$-median and to the flag mean. We find that using FlagIRLS to compute the flag median converges in $4$ iterations on a synthetic dataset. We also see that Grassmannian LBG with a codebook size of $20$ and using the flag median produces at least a $10\%$ improvement in cluster purity over Grassmannian LBG using the flag mean or $\ell_2$-median on the Mind's Eye dataset.
\end{abstract}



\section{Introduction}
The mean and median are basic methods for calculating central prototypes from a probability distribution. The median is commonly more robust to outliers than the mean. 
Generalizations of such prototypes to Euclidean space can be formulated as a solution to an optimization problem. Suppose we have a set of points in Euclidean space, $\mathcal{X} = \{\mathbf{x}_i\}_{i=1}^p \subset \R^n$ which we would like to
represent as a prototype $\mathbf{y}$ via
solving
\begin{equation}\label{eq:euclopt}
\arg \min_{\mathbf{y} \in A} \sum_{i=1}^p \|\mathbf{x}_i - \mathbf{y}\|^q_2.
\end{equation}
The solution to~\eqref{eq:euclopt} for $A=\R^n$ and $q=2$ is called the centroid, which may be viewed as the generalization of the mean. In fact, the centroid is the component-wise mean of the vectors in $\mathcal{X}$, $\sum_{i=1}^p \bx_i/p$.  Generalizations of the median involve solving~\eqref{eq:euclopt} when $q=1$. When $A=\cX$, the solution is called the medoid, while when $A=\R^n$, the solution is called the geometric median. The geometric median inherits the robustness to outliers from the median without being required to be a point in the dataset; however, it is not as straightforward to compute as a centroid since that calculation is not simply a least squares problem. An iterative algorithm for approximating a geometric median is the Weiszfeld algorithm~\cite{weiszfeld1937point}; each iteration of this algorithm is a weighted centroid problem. Thus, the Weiszfeld algorithm falls into a class known as Iteratively Reweighted Least Squares algorithms (IRLS). These Euclidean prototypes are used as a statistic for a dataset and in common machine learning algorithms like $k$-means and nearest centroid classification.

Not all datasets are best represented using points in Euclidean space. Specifically, image or video datasets are sometimes better represented using subspaces, i.e., as points on a Grassmannian.  For example, the smallest principal angle between two subspaces
has proven powerful for modeling illumination
spaces~\cite{beveridge2008principal}.
Hyperspectral  data may fail to be linearly separable in Euclidean space but separate
linearly on the Grassmannian~\cite{chepushtanova2017sparse}.
Hence, it is potentially useful to find versions of the Euclidean prototypes on the Grassmannian. A logical generalization of prototypes from Euclidean space to the Grassmannian is to replace the Euclidean $2$-norm in the optimization problems for the centroid, medoid and geometric median with a distance or dissimilarity between subspaces. The centroid is generalized using the geodesic distance in \cite{karcher1977riemannian} and the chordal distance in \cite{draper2014flag}. To the extent of our research, we have have not found a generalization of the medoid. However, the geometric median has been generalized using the geodesic distance and is called the $\ell_2$-median in \cite{aftab2014generalized, fletcher2009geometric}. Prototypes like these have been used alone as a method to classify emotion in images \cite{zhang2018grassmannian}, as a step in a k-means type algorithm \cite{marrinan2014finding} and in feature extraction \cite{fletcher2009geometric}.

Popular machine learning techniques, like dictionary learning, have been adapted to Riemannian manifolds \cite{xie2012dictionary}. Jayasumana et.\ al.\ consider learning on the Grassmannian (and Riemannian manifolds in general) with RBF kernels and advocate for chordal distance (sometimes referred to as the projection norm) kernels on the Grassmannian because chordal distance generates a positive definite Gaussian kernel \cite{jayasumana2015kernel}. More recently, Cherian et.\ al.\ use kernalized Grassmannian pooling for activity recognition \cite{cherian2018non}. Methods for Riemannian optimization like Riemannian SVRG have gained popularity alongside this surge of interest in Riemannian learning \cite{zhang2016riemannian}.  Even more uses for subspaces in computer vision and machine learning can be found in, e.g.,  \cite{aftab2014generalized,bates2015max,draper2014flag,edelmangeometry,marks2012mean,marrinan2014finding,zhang2018grassmannian}.

In this paper we propose the flag median, a prototype which is a generalization of the geometric median to the Grassmannian using the chordal distance. We solve the flag median optimization problem using the novel FlagIRLS algorithm. The FlagIRLS is an IRLS algorithm on the Grassmannian that solves a weighted flag mean problem at each iteration similar to the way an iteration of the Weiszfeld algorithm solves a weighted centroid problem. We conduct experiments with the flag median, $\ell_2$-median and the flag mean on synthetic datasets, the MNIST handwritten digits dataset  \cite{deng2012mnist}, the DARPA (Defense Advanced Research Projects Agency) Mind's Eye dataset used in \cite{marrinan2014finding} and the UCF YouTube action dataset \cite{liu2009recognizing}. In these examples we find that the FlagIRLS algorithm tends to converge quickly.  We show that the flag 
median appears to be more robust
to outliers than the flag mean and
$\ell_2$-median,
and produces the highest cluster purities in the LBG algorithm \cite{linde1980algorithm}.

\section{Background}\label{sec:background}

\subsection{Introduction to the Grassmannian}
For the purposes of this paper, the Grassmannian manifold (a.k.a. "the Grassmannian"), denoted $\Gr(k,n)$, is the manifold whose points correspond to the $k$ dimensional subspaces of  $\R^n$. We will represent a point in $\Gr(k,n)$ using a tall $n \times k$ real matrix $\X$ with orthonormal columns. The point on $\Gr(k,n)$ determined by $\X$ is the column space of $\X$ and is denoted $[\X]$. Thus if $\X$ and $\Y$ have the same column space then they determine the same point $[\X]=[\Y]$ on $\Gr(k,n)$.

In order to allow more flexibility in our generalization of optimization problems to subspaces, we work with points that are not all necessarily on the same Grassmannian manifold but are in the same ambient space. Suppose we have a set of subspaces of $n$-dimensional space, $\{[\X_1],[\X_2], \dots, [\X_p]\}$, where $[\X_i] \in \Gr(k_i, n)$. We want to find an $r$-dimensional subspace of $\R^n$, $[\Y^*] \in \Gr(r,n)$, that is in some sense the center of these points, i.e., that $[\Y^*]$ is a solution to 
\begin{equation}
    \arg \min_{[\Y] \in \Gr(r,n)} \sum_{i=1}^p d([\X_i],[\Y])
\end{equation}
where $d$ measures dissimilarity between its arguments.


Principal angles between subspaces are a common dissimilarity measure that is invariant to orthogonal transformations \cite{bjorck1973numerical,draper2014flag}. Take $[\X], [\Y] \in \Gr(k,n)$. The $i$th smallest principal angle between $[\X]$ and $[\Y]$, $\theta_i ([\X], [\Y]) \in [0, \pi/2]$ is defined as the solution to~\eqref{eq: principal angle problem} \cite{bjorck1973numerical}.
\begin{align}\label{eq: principal angle problem}
\begin{aligned}
    \cos \theta_i ([\X], [\Y]) &= \max_{\x \in [\X]} \: \: \max_{\y \in [\Y]} \: \: \x^T \y = \x_i^T \y_i\\
    \text{Subject to } &\x^T\x = \y^T \y = 1\\
    & \x^T \x_j = \y^T \y_j = 0 \text{ for } j=1,2,\dots, i-1
\end{aligned}
\end{align}
Now let $\theta([\X], [\Y]) \in \R^k$ be the vector of principal angles between $[\X]$ and $[\Y]$. The geodesic distance on $\Gr(k,n)$ is $\| \theta([\X], [\Y]) \|_2$ and the chordal distance on $\Gr(k,n)$ is $\| \sin( \theta([\X], [\Y]) ) \|_2$ \cite{conway1996packing}. We can calculate these quantities when $[\X] \in \Gr(k,n)$, $[\Y] \in \Gr(r,n)$ where $k \neq r$ by setting the last $\max(k,r) - \min(k,r)$ entries of $\theta([\X], [\Y]) \in \R^{\max(k,r)}$ to $0$.

\subsection{Geodesic Distance Prototypes}

The Euclidean mean and geometric median have been translated to the Grassmannian using the geodesic distances. The mean on the Grassmannian using geodesic distance (the solution to~\eqref{eq:geodes} for $q=2$) is called the Karcher mean and the geometric median using geodesic distance (the solution to~\eqref{eq:geodes} for $q=1$) is called the $\ell_2$-median. 
\begin{equation}\label{eq:geodes}
    \arg \min_{[\Y] \in \Gr(r,n)} \sum_{i=1}^p \| \theta ([\X_i],[\Y])\|_2^q.
\end{equation}
The Karcher mean and the $\ell_2$-median are only computable in the case where all subspaces are of equal dimensions (e.g., $r = k_1 = k_2 \cdots = k_p$) and \cite{marrinan2014finding} use examples to show that the available algorithms to compute these prototypes are slow. 
The most common algorithm for finding the solution to the Karcher mean was discovered by Karcher \cite{karcher1977riemannian} and  Fletcher et.\ al.\ \cite{fletcher2009geometric} show we can find the $\ell_2$-median using a Weiszfeld-type algorithm. \cite{marrinan2014finding} show that the Karcher mean is not only slow to compute, but also produces lower cluster purities than the $\ell_2$-median in their LBG clustering example so we choose not to use the Karcher mean as a prototype in our experiments (Section \ref{sec:experiments}).

 For context, the Weiszfeld algorithm for vectors in $\R^n$ is stated in Algorithm~\ref{alg:weis_algorithm}. 
  \\
\begin{algorithm}[ht]
\SetAlgoLined
 \KwData{ $\{\x_i\}_{i=1}^p \subset \R^n$}
 \KwResult{ The geometric median $\y \in \R^n$}
 \While{not converged}{
  $w_i = \frac{p}{\| \x_i - \y \|_2} \left( \sum_{k=1}^p \frac{1}{\| \x_k - \y \|_2} \right)^{-1}$;\\
  $\y \leftarrow \sum_{i=1}^p\frac{w_i \x_i}{p}$;
 }
 \caption{Weiszfeld Algorithm in $\R^n$}\label{alg:weis_algorithm}
\end{algorithm}
\\
Note that each iteration of the Weiszfeld algorithm (Algorithm \ref{alg:weis_algorithm}) is the solution to the least squares problem~\eqref{eq:euclopt} (with $A=\R^n$ and $q=2$) for the weighted vectors $w_i \x_i$.
The weights, $w_i$, come from the fact that the geometric median $\y$ satisfies (presuming $\y \notin \cX$)
\[
\y = \left(\sum_{i=1}^p \frac{\x_i}{\| \x_i - \y \|_2} \right) \Big/ \left( \sum_{k=1}^p \frac{1}{\| \x_k - \y \|_2} \right).
\]

Fletcher et.\ al.\ solve~\eqref{eq:geodes} for $q=1$ by generalizing this approach to Riemannian manifolds. 

 In Section \ref{sec:experiments}, we use the unweighted Weiszfeld-type algorithm from Fletcher et.\ al.\ with geodesic distance to calculate the $\ell_2$-median on the Grassmannian. Let $d$ be the maximum distance between points in the dataset and let $\delta$ be the convergence parameter. We define $N_{d,\delta}$ as the number of iterations of one run of our implementation. The complexity of our implementation of this algorithm in Section \ref{sec:experiments} is ${O}\left(n p k^2N_{d,\delta}\right)$. 

\subsection{The Flag Mean Prototype}
Draper et.\ al.\ \cite{draper2014flag} present the flag mean as an average of subspaces of different dimensions using the squared chordal distance. The optimization problem for the flag mean is 
\begin{equation}\label{eq:flag_mean}
\arg \min_{[\Y] \in \Gr(r,n)} \sum_{i=1}^p \| \sin(\theta ([\X_i],[\Y]))\|_2^2.
\end{equation}
This flag mean determines not only a point on $\Gr(r,n)$, it determines a point on various flag manifolds. A flag manifold is a manifold whose points represent a flag of subspaces  $[\mathbf{S}_1] \subset [\mathbf{S}_2] \subset \dots \subset [\mathbf{S}_r] = \mathbb R^n$. If we let $s_i = \text{dim}([\mathbf{S}_i])$, then we say the flag is of type  ${s_1,s_2,\dots,s_r}$. For more details on flag manifolds, see \cite{monk1959geometry}.

We will refer to a flag mean in this paper as the point on $Gr(r,n)$ determined by the flag. Let $[\Y]$ be the flag mean of $\{[\X_i]\}_{i=1}^k$. Let $\y_i$ be the $i$th column of $\Y$, the orthonormal matrix representation of $[\Y]$. Then the $r^{th}$ ``real'' flag mean is the point on the flag manifold of type $\{1,2,\dots, r, n\}$ defined as in~\eqref{eq:point_on_flag}.

\begin{equation}\label{eq:point_on_flag}
    \llbracket \Y \rrbracket = \text{span}\{\y_1\} \subset \text{span}\{\y_1,\y_2\} \subset \dots \subset \text{span}\{\y_1,\y_2, \dots , \y_r\} \subset \mathbb R^n
\end{equation}

The point on $Gr(r,n)$ determined by the flag is $\text{span}\{\y_1,\y_2, \dots , \y_r\} \subset \mathbb R^n$.

Draper et.\ al.\ ~\cite{draper2014flag} show that we can calculate the flag mean by utilizing the singular value decomposition (SVD) of the matrix $[ \X_1, \X_2, \dots , \X_p ]$. The flag mean, as a point on $Gr(r,n)$, is the span of the $r$ left singular vectors of the  corresponding to the $r$ largest singular values.

The complexity of this algorithm is $O\left(n \left(\sum_{i=1}^{p}k_i\right)^2\right)$. Marks \cite{marks2012mean} suggests calculating weighted flag means in his dissertation. This weighted flag mean calculation will be used as an iteration of the FlagIRLS algorithm introduced in Section \ref{sec:flagirls}.

\section{Flag Median}
The translation of the geometric median to the Grassmannian using chordal distance is called the flag median. The optimization problem for this novel prototype is in~\eqref{eq:Flag Median}.

\begin{equation}\label{eq:Flag Median}
    \arg \underset{[\Y] \in Gr(r,n)}{\min} \sum_{i=1}^p \|\sin \theta([\X_i], [\Y])\|_2
\end{equation}
We call this the flag median since, using FlagIRLS, $[\Y]$ actually is a flag of subspaces rather than a single $r$ dimensional subspace of $n$ dimensional space. This flag median is indeed a median (similar to the geometric median) because it minimizes the chordal distance rather than the squared chordal distance problem in~\eqref{eq:flag_mean} that is solved by the flag mean.

\subsection{Derivation}\label{sec:derivations}
In this section we show that the FlagIRLS algorithm can be used to approximate the flag median. The algorithm derived in this section revolves around weighted flag means of $\{[\X_i]\}_{i=1}^p$. For the rest of this paper we will denote the weight of the subspace $[\X_i]$ as $w_i$. 

Notice that the flag median optimization problem in~\eqref{eq:Flag Median} involves a sum of two norms of the vector of sines of principal angles and the flag mean optimization problem in~\eqref{eq:flag_mean} involves squared two norms of the same vector. So, in other words, we are deriving an algorithm, analogous to Weiszfeld and IRLS in the Euclidean setting, that approximates solutions to the $2$-norm problem by iteratively solving squared $2$-norm problems. So the FlagIRLS algorithm \ref{alg:main_algorithm} provides an ``iterative reweighted least squares'' method for approximating the flag median.

 Let us begin by translating the flag median problem from~\eqref{eq:Flag Median} to an optimization problem over matrices with orthonormal columns. The eigenvalues of $\Y^T \X_i \X_i^T  \Y$ are the entries in the vector $\cos^2( \theta([\X_i], [\Y]))$. Using properties of trace we can show $\tr(\Y^T \X_i \X_i^T  \Y) = \sum_{j=1}^{m_i} \cos^2 \theta_j([\X_i], [\Y])$ \cite{bjorck1973numerical}. This allows us to rewrite the flag median problem from~\eqref{eq:Flag Median} as the matrix optimization problem in~\eqref{eq:sin_matrix} where $m_i = \min(r,k_i)$.  
\begin{equation}\label{eq:sin_matrix}
    \min_{\substack{\Y \in \R^{n \times r} \\ \Y^T\Y = I}} \sum_{i=1}^p \left( m_i-\tr(\Y^T \X_i \X_i^T\Y)\right)^{1/2}
\end{equation}
We formulate a Lagrangian from this problem using $\Lambda$ as a symmetric matrix of Lagrange multipliers with entries $\lambda_{ij}$ in~\eqref{eq:sin_lagrange}.

\begin{align}\label{eq:sin_lagrange}
\begin{aligned}
\mathcal{L}(\Y,\Lambda) = &\sum_{i=1}^p \left(m_i-\tr(\Y^T \X_i \X_i^T \Y)\right)^{1/2} \\
&- \langle \Lambda, \Y^T\Y - I \rangle
\end{aligned}
\end{align}
We then calculate~\eqref{eq:sin_algebra} the Lagrangian with respect to the $j$th column of $\Y$, namely $\y_j$, and set it equal to $0$. 
 
\begin{equation}\label{eq:sin_algebra}
    \y_j^T \sum_{i=1}^p \frac{-1}{\left(m_i-\tr(\Y^T \X_i \X_i^T \Y)\right)^{1/2}} \X_i \X_i^T \y_j = 2\lambda_{jj}
\end{equation}
Now define the matrix $\X$
\begin{equation}
\label{eq:big mat}    
\X = \left[  w_1 \X_1 , w_2 \X_2 , \cdots , w_p \X_p \right]
\end{equation}
where $ w_i = \left(\frac{1}{m_i-\tr(\Y^T \X_i \X_i^T \Y)}\right)^{1/4}.$

Combining the information in \eqref{eq:sin_algebra} and the matrix $\X$ in \eqref{eq:big mat}, we see that $\Y$ must be $r$ left singular vectors of $\X$ when $[\Y]$ to be the flag median of $\{[\X_i]\}_{i=1}^p$. 

Now let us consider an iterative algorithm with the $j$th iteration of the form $\Y_{j+1} = \text{Flag Mean}\left( \left \{w_i^{(j)}  \X_i\right \}_{i=1}^p \right)$ where
$w_i^{(j)} = \left(\frac{1}{m_i-\tr(\Y_j^T \X_i \X_i^T \Y_j)}\right)^{1/4}$.
This algorithm will be formalized in Section \ref{sec:flagirls}. For this type of algorithm, we desire $\Y_{j+1}$ to be an approximation of the flag median of the dataset $\{\X_i\}_{i=1}^p$. We will now show that the columns of $\Y_{j+1}$ should be chosen as the left singular vectors of $\X$ associated with the $r$ largest singular values and therefore the update using the flag mean of $\left \{w_i^{(j)}  \X_i \right \}_{i=1}^p$ is the correct update choice. 

Let $\U^*$ be some matrix consisting of $r$ left singular vectors of $\X$. To determine $\mathbf{U}^*$ where
\begin{equation}\label{eq:sine_reiterated}
 \mathbf{U}^* = \arg \min   \sum_{i=1}^p \left( q - \tr (\mathbf{U}^T \X_i \X_i^T \mathbf{U}) \right)^{1/2}.
\end{equation} 
we solve the optimization problem
\begin{equation}\label{eq:sine_opt}
\begin{aligned}
    &\max_{\substack{ \mathbf{U} \in \R^{n \times r} \\ \mathbf{U}^T \mathbf{U} = \I}}  \tr \left( \mathbf{U}^T \X \X^T \mathbf{U} \right) \\
\end{aligned}
\end{equation}
which requires the columns of $\mathbf{U}^*$ to be the left singular vectors of
$\X$
associated with the largest singular values.  So we take our update to be $\Y_{j+1} = \mathbf{U}^*$.

\subsection{The FlagIRLS Algorithm}\label{sec:flagirls}
We use an iteratively reweighted least squared flag mean algorithm structure to solve~\eqref{eq:Flag Median}. We will call the weight for subspace $\X_i$, $w_i$. A concern with these weights arises when the denominator of a weight is zero. For the flag median objective function, the denominator is zero when $[\Y]$ is a subspace of $[\X]$ or vise versa. 
To avoid singularities, we added a small quantity $\epsilon$ to the denominator. The $w_i$ for the flag median problem is in~\eqref{eq:weights}
\begin{equation}\label{eq:weights}
     w_i = \left( \frac{1}{m_i-\tr(\Y^T \X_i \X_i^T \Y)+\epsilon}\right)^{1/4}
\end{equation}
We use these weights, along with the flag mean, in the FlagIRLS algorithm as described in Algorithm \ref{alg:main_algorithm}.

\begin{algorithm}[ht]
\SetAlgoLined
 \KwIn{ A set of orthonormal subspace representatives $\{\X_i\}_{i=1}^p$ for $\{[\X_i] \in \Gr(k_i, n)\}_{i=1}^p$}
 \KwOut{ An orthonormal subspace representative $\Y$ for the flag median $[\Y] \in \Gr(r,n)$}
 \While{not converged}{
  assign each $w_i$\;
  $\X \leftarrow \left [ w_1 \X_1 | w_2 \X_2 | \cdots | w_p \X_p \right]$\;
  $\U \Sigma  \V ^T = \X$ \%calculate the SVD \; 
  $\Y \leftarrow \U [:,1:r]$ \%first $r$ columns of $\U$;
 }
 \caption{The FlagIRLS algorithm. See~\eqref{eq:weights} for the algorithm weights. We assume the columns of $\U$ are sorted from the left singular vector associated with the smallest to the largest singular values of $\X$.} \label{alg:main_algorithm}
\end{algorithm}

An important note is that FlagIRLS is an iterative weighted flag mean algorithm so the outputs of this algorithm come from the left singular vectors of $\X$. We take $r$ singular vectors associated with the $r$ largest singular values of $\X$, i.e.,  the first $r$ columns of $\U$. However, there are $n$ columns of $\U$, so FlagIRLS actually outputs a flag of subspaces
   $ [\U [:,1]] \subset [\U [:,:2]] \subset \cdots \subset [\U [:,:n]].$
This flag is used to distinguish between different prototypes in Section \ref{sec:mnist_experiments} with MNIST digits.

\section{Limitations}

The main limitation of calculating the flag median is the speed of FlagIRLS. This requires that we take the thin SVD of $\X \in \R^{n \times pk}$ every iteration in FlagIRLS. The complexity of the FlagIRLS algorithm is the complexity of the flag mean times the number of iterations of the algorithm, i.e.,  $O\left(n N_{\delta} \left(\sum_{i=1}^{p}k_i\right)^2\right)$ where $N_{\delta}$ is the number of iterations and $\delta$ is the convergence parameter.


Another current limitation of this work is the lack of proven mathematical guarantees for the flag median and the FlagIRLS algorithm. Although, for all our examples, FlagIRLS converges to a local minimum of the flag median problem, we have not worked out the mathematical theory to find the conditions where FlagIRLS converges. We also still need to determine the conditions where an iteration of FlagIRLS is a contraction mapping. On a larger scale, given a dataset of subspaces of $\R^n$, we have yet to determine where the flag median problem is convex. Section \ref{sec:derivations} shows that FlagIRLS is a logical algorithm for finding the flag median, but further development of the mathematical theory would give us more intuition about which datasets are good for FlagIRLS, how to initialize FlagIRLS and overall provide the user with
a better understand of rates of convergence.
Currently, the FlagIRLS algorithm is run with a number of different initializations to verify convergence.






\section{Experiments}\label{sec:experiments}
In this section we carry out experiments with synthetic data, the MNIST handwritten digits dataset \cite{deng2012mnist}, the Mind's Eye dataset \cite{marrinan2014finding} and the UCF YouTube action dataset \cite{liu2009recognizing}. The goal is to compare the flag median to the flag mean and the $\ell_2$-median and establish the efficiency of FlagIRLS. For all of this section we use FlagIRLS to compute the flag median and the Weiszfeld-type algorithm from \cite{fletcher2009geometric} for the $\ell_2$-median.

 The convergence criteria for our implementation of FlagIRLS is as follows. We terminate the algorithm when objective function values of consecutive iterates of FlagIRLS are less than $\delta = 10^{-11}$, or if the $i$th iteration resulted in an increasing objective function value. In the former case, we output the $(i-1)$st iterate. For our weights in all examples we run the FlagIRLS algorithm with $\epsilon = 10^{-7}$. 

The convergence criteria of our implementation of the Weiszfeld-type algorithm from \cite{fletcher2009geometric} to calculate the $\ell_2$-median is similar to the Flag IRLS convergence criteria. 
We terminate the algorithm when when objective function values of consecutive iterates are less than $\delta =10^{-11} $.

Both FlagIRLS and the Weiszfeld-type algorithm are terminated when we have exceeded $1000$ iterations. FlagIRLS never exceeds $1000$ iterations in our examples. 

\subsection{Synthetic Data}

We begin with two experiments on a dataset consisting of $10$ points from $\Gr(3,20)$ and $10$ points from $\Gr(5,20)$. A representative for a point on $\Gr(k,n)$ is sampled in two steps. The first step is to sample an $n \times k$ matrix from a uniform distribution on $[-.5, .5)$, $\mathcal{U} [-.5, .5)$. We then do the QR decomposition of this matrix to get a point on $\Gr(k,n)$. We perform two experiments on this dataset: The first experiment verifies convergence of FlagIRLS, and the second experiment compares the convergence rate of FlagIRLS to Grassmannian gradient descent.

For the first experiment, we run 100 trials of FlagIRLS with different random initializations. For each of these trials, we verify that we have converged by checking 100 points near the FlagIRLS algorithm output. Given one algorithm output, $[\X] \in \Gr(3,20)$, we sample the entries of $\Y \in \mathbb{R}^{20 \times 3}$ from $\mathcal{U} [-0.5, 0.5) $ and check the objective function value at the first $3$ columns of $\Q$ where $\Q$ comes from the QR decomposition of the matrix $\X + 0.00001 \Y$. We call these points ``test points'' for the algorithm output. We say the FlagIRLS algorithm for flag median converged when all the objective function values of the test points are less than or equal to the objective function value for the algorithm output. In this experiment, we find that $100 \%$ of the FlagIRLS trials converge. 

\begin{figure}[ht]
    \centering
        \includegraphics[width=.7\textwidth]{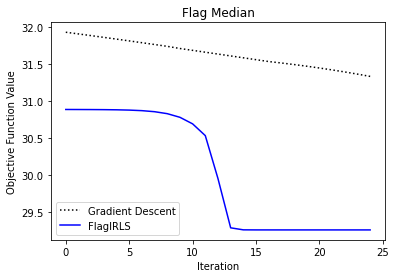}
    \caption[]{The mean objective function values over $100$ trials with different random initializations. FlagIRLS converges in fewer iterations than gradient descent for the flag median problem for the synthetic dataset.}
    \label{fig:convergence examples}
\end{figure}

We now show an example with the same dataset where we run FlagIRLS and Grassmannian gradient with $100$ random intializations to compute the flag median. The results of this experiment are in Figure \ref{fig:convergence examples}. For this example, Grassmannian gradient descent is implemented with a step size of $0.01$. We  find that FlagIRLS converges in fewer iterations than Grassmannian gradient descent for the flag median problem. 

For our next example, we use a dataset of $200$ points on $\Gr(6,100)$. The points are sampled by first fixing a ``center'' point for the dataset, $[\X_*]$. We do this by taking a random $100 \times 6$ matrix with entries from $\mathcal{U} [-.5, .5)$. We then take $\X_*$ as the first $6$ columns of $\Q$ from the QR decomposition of this random matrix. The $200$ points in the dataset are now calculated via the following steps. For each point, we generate $\Z$ by sampling a random $100 \times 6$ matrix with entries sampled from $\mathcal{U} [-.5, .5)$ and scaling it by $0.01$. We then take the point determined by the first $6$ columns of $\Q$ from the QR decomposition of $\X_* + \Z$.

We then run our FlagIRLS and Weiszfeld-type algorithm implementations with 
$20$ random initializations to calculate the flag median and the $\ell_2$-median respectively. For the random initializations, we initialize FlagIRLS and the Weiszfeld-type algorithm at the same point. The results of this experiment are in Table \ref{tab:FlagIRLS_vs_Weiszfeld-type}. We terminate the Weiszfeld-type algorithm after $1000$ iterations regardless of convergence. So perhaps, many of the high iteration runs of Weiszfeld still did not converge even after $1000$ iterations.
\begin{table}[]
    \centering
    \begin{tabular}{c||c}
         Algorithm/ Initialization &  Mean Iterations\\
         \hline
         \hline
        FlagIRLS/random&  $4.55 \pm 0.50$\\
        \hline
        Weiszfeld-Type/ datapoint &  $795.50 \pm 227.40$\\
        \hline
        Weiszfeld-Type/ randomly & $968.80 \pm 94.49$\\
    \end{tabular}
    \caption{The mean number of iterations until convergence of $20$ random initalizations of FlagIRLS and the Weiszfeld-type algorithm on a dataset of $200$ points on $\Gr(6,100)$. FlagIRLS is converges in far fewer iterations than the Weiszfeld-type algorithm and also sports a much lower standard deviation in the number of iterations.}
    \label{tab:FlagIRLS_vs_Weiszfeld-type}
\end{table}





Now we will use a dataset that consists of $200$ points on $\Gr(3,20)$. These points consist of a cluster of $180$ points centered around the subspace $[\X_*]$ and $20$ outlier points. The points from the $180$-point cluster are sampled by the following process. We calculate a fixed ``center'' point for the dataset, $[\X_*]$, by taking a $20 \times 3$ matrix with entries from $\mathcal{U} [-.5, .5)$. We then take $\X_*$ as the first $3$ columns of $\Q$ from the QR decomposition of this random matrix. We then calculate the points in the cluster via the following steps. The first step is to generate $\Z$ by sampling a random $20 \times 3$ matrix with entries sampled from $\mathcal{U} [-.5, .5)$ and scaling it by $0.01$. The second step generates one point in the 180 point cluster as the first 3 columns of $\Q$ from the QR decomposition of $\X_* + \Z$. A point from the set of outlier 20 points is the first $3$ columns of the QR decomposition of a random $20 \times 3$ matrix with entries sampled from $\mathcal{U} [-.5, .5)$.


Table \ref{tab:c_dists_center} shows the results of calculating the flag median, $\ell_2$-median and flag mean of this dataset and then computing the chordal distance between $[\X_*]$ and the three different prototypes. Notice the flag median is the least affected by the outliers, the $\ell_2$-median is twice as affected and the flag mean is ten times more affected by the outliers.

\begin{table}[ht]
    \centering
    \begin{tabular}{c|c}
    Algorithm & Chordal Distance\\
    \hline
    Flag Median    &  0.0017\\
    $\ell_2$-median    &  0.0022\\
    Flag Mean    &  0.0128\\
    \end{tabular}
    \caption{The chordal distance between the algorithm result and $[\X^*]$.}
    \label{tab:c_dists_center}
\end{table}
\textit{Note: for Table \ref{tab:c_dists_center}, FlagIRLS converges to the flag median in one iteration.}

\subsection{MNIST Handwritten Digits dataset}\label{sec:mnist_experiments}

 The MNIST digits dataset is a set of $28 \times 28$ single band images of handwritten digits \cite{deng2012mnist}. We represent an MNIST handwritten digit using an element of $\Gr(1,784)$ to by taking one image, vectorizing it, then dividing the resulting vector by its norm. 

\begin{figure}[ht]
    \centering
    \includegraphics[width=.7\textwidth]{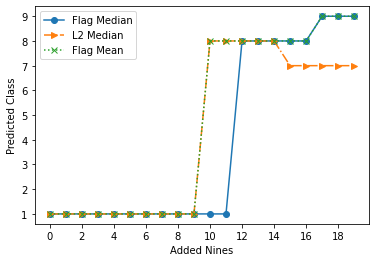}
    \caption[]{The neural network predicted class of the prototype for the dataset with $20$ examples of $1$'s and $i$ examples of $9$'s with $i=0,1,2,\dots,19$.}
    \label{fig:MNIST_classifier}
\end{figure}

For our first example, we see how the flag median, $\ell_2$-median and flag mean prototypes are classified by a MNIST-trained 3-layer neural network. This trained neural network classifier has a $97\%$ test accuracy on the MNIST test dataset. We generate our datasets for this experiment by to taking 20 examples of the digit $1$ and $i$ examples of the digit $9$ from the MNIST training dataset. We let $i=0,1,2,3, \dots, 19$ and this results in $20$ datasets. For each of these datasets, we calculate the flag median, $\ell_2$-median and flag mean, then predict the class of each of these prototypes by passing each through the neural network classifier. 
In Figure \ref{fig:MNIST_classifier} we plot the predicted class of each prototype for each dataset by the trained neural network. For this figure, we choose to use the random initialization that resulted in the best predictions of the $\ell_2$-median.

The $\ell_2$-median and flag mean are misclassified with $i=9$ added examples of $9$'s whereas the flag median is still classified correctly for $i=10$ and $i=11$ added examples $9$'s. Therefore the flag median is the most robust prototype to outliers in this experiment. The common misclassification as $8$ is likely due to the fact that the $1$'s tend to be at an angle, so when averaged, they tend to look like fuzzy $8$'s, especially when some $9$'s have been introduced to the dataset.
Also, the $\ell_2$-median of a dataset of $20$ $1$'s with $15$ to $19$ $9$ digits is misclassified as a $7$. This is likely a result of the different angled $1$'s and the introduction of the examples of $9$'s adding the top of the digit $7$.

\begin{figure}[ht]
    \centering
    \begin{minipage}[b]{.45\textwidth}
    \includegraphics[width=\textwidth]{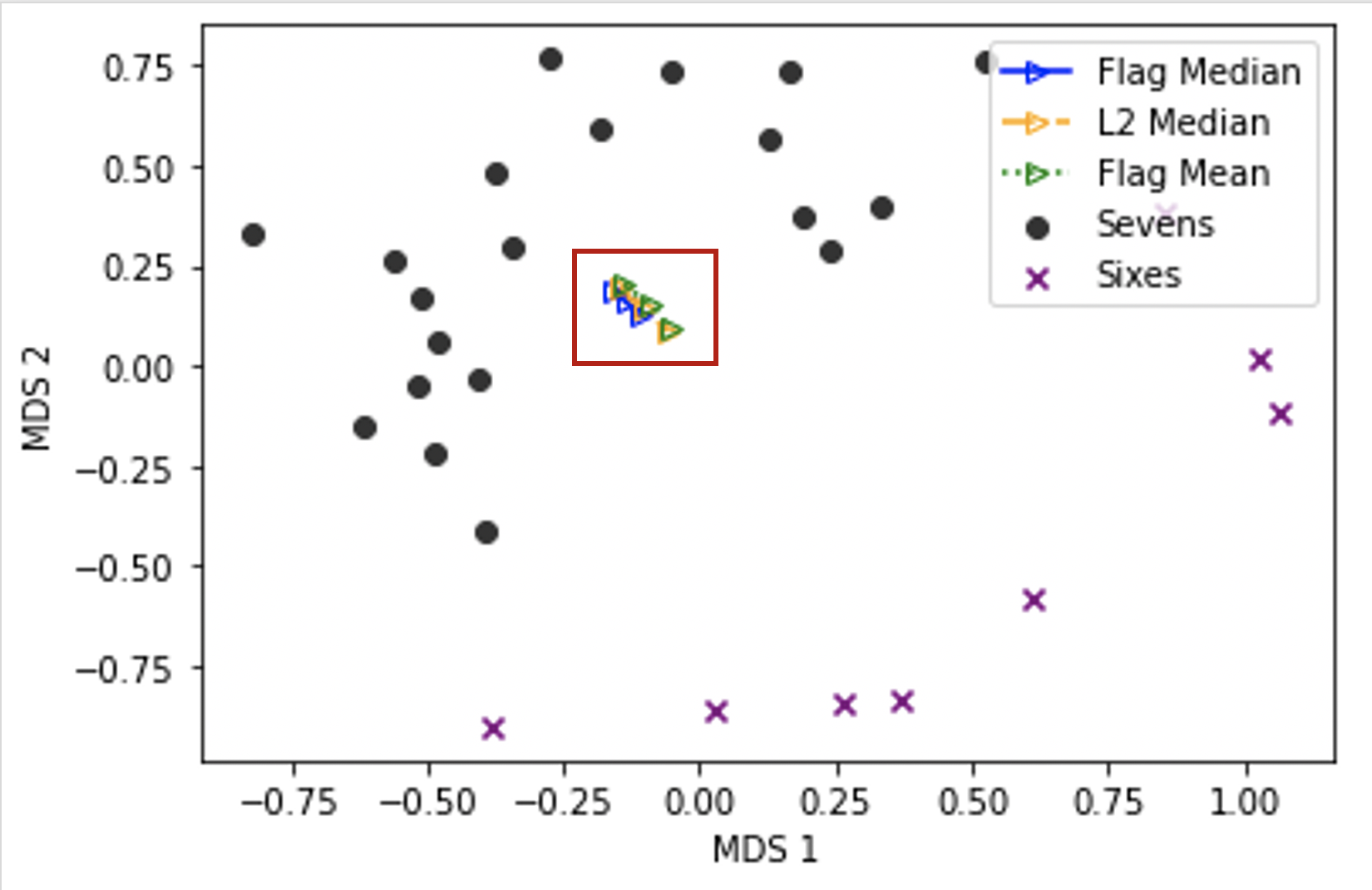}
    \end{minipage}\hfill
    \begin{minipage}[b]{.45\textwidth}
    \includegraphics[width=\textwidth]{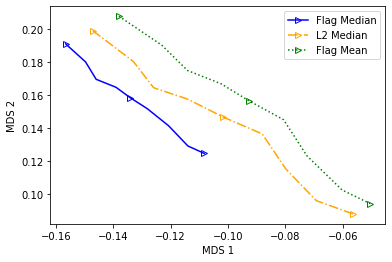}
    \end{minipage}
    \caption[]{MDS embedding of flag median, $\ell_2$-median and flag mean with points as one dimensional subspaces. We have $20$ examples of $7$'s and $i$ examples of $6$'s. Each triangle represents a prototype for $i=0,4,8$. The furthest left triangle is the prototype for the dataset with $i=0$ examples of $6$'s and the furthest right triangle is the prototype for the dataset with $i=8$ examples of $6$'s. The lower image is a zoomed in version of the interior of the red box in the upper image to clarify the difference between the exemplars.}
    \label{fig:1MDS}
\end{figure}

Now we use Multi Dimensional Scaling (MDS) \cite{kruskal1978multidimensional} to visualize the movement of the prototypes of an MNIST dataset that is poisoned with outliers. For this experiment, we use $20$ examples of $7$'s and $i=0,2,4,6,8$ examples of $6$'s. This results in $5$ different subspace datsets formed from examples from the MNIST training dataset. We then calculate the flag median, $\ell_2$-median and the flag mean. We generate a distance matrix for all the examples of $6$'s and $7$'s along with the exemplars from each dataset using the geodesic distance and pass the distance matrix through a MDS algorithm to visualize relationships between these subspaces in two dimensions. This example is in Figure \ref{fig:1MDS}.

Notice that the flag mean is moving the most as we add examples of $6$'s and the $\ell_2$-median is moving similarly to the flag mean. The flag median moves substantially less than the other prototypes and therefore is the least affected prototype by the added examples of $6$'s.

\begin{figure}[ht]
    \centering
    \includegraphics[width=.9\textwidth]{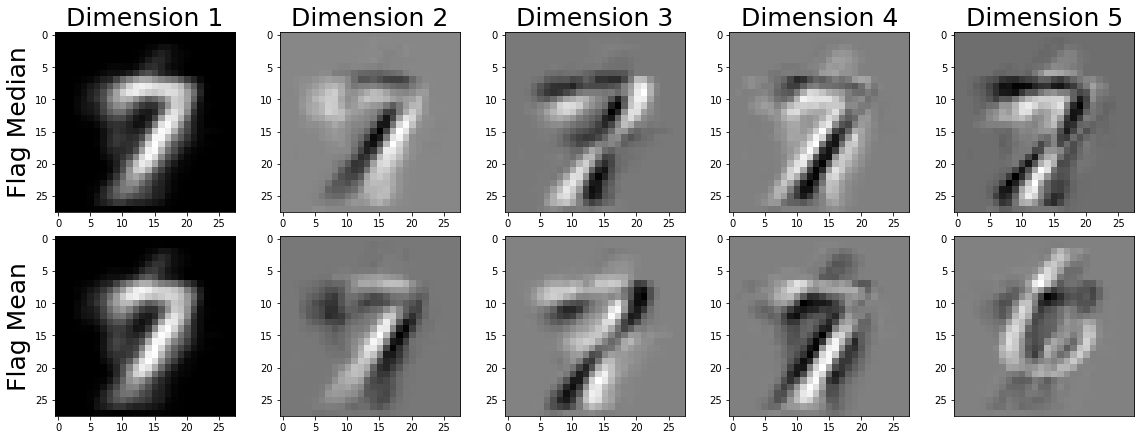}
    \caption[]{Each column of the matrix representative for flag median and flag mean on the dataset with $20$ examples of $7$'s and $i=8$ examples of $6$'s.}
    \label{fig:high_dim}
\end{figure}

We now compute the $r=5$-dimensional flag median and flag mean of a dataset with 20 examples of $7$'s with $i=8$ $6$'s. We plot each of the reshaped columns of the matrix representative of these prototypes in Figure \ref{fig:high_dim}. Notice that the flag mean is more affected by examples of $6$'s than the flag median. This is particularly noticeable in the final column (dimension $5$) where there is a clear $6$ in the image for flag mean whereas the $6$ is not clear in the flag median.

\subsection{Mind's Eye dataset}\label{sec:mindseye_experiments}
The Mind's Eye dataset is a set of grey-scale outdoor video clips that are centered on moving objects (mainly humans) and have a subtracted background. Each video clip consists of 48 frames, each rescaled to a size of $32 \times 32$ pixels. We use the preprocessed data from the k-means experiment from Marrinan et.\ al.\ \cite{marrinan2014finding}. These data and the scripts for the preprocessing can be accessed at \url{https://www.cs.colostate.edu/~vision/summet}. There are 77 labels of the video clips for the action of the centered object in the video. A video clip is represented on $\Gr(48,1024)$ by the span of the $1024 \times 48$ matrix formed by vectorizing and horizontally stacking each frame.

\begin{figure}[ht]
    \centering
    \includegraphics[width=.8\textwidth]{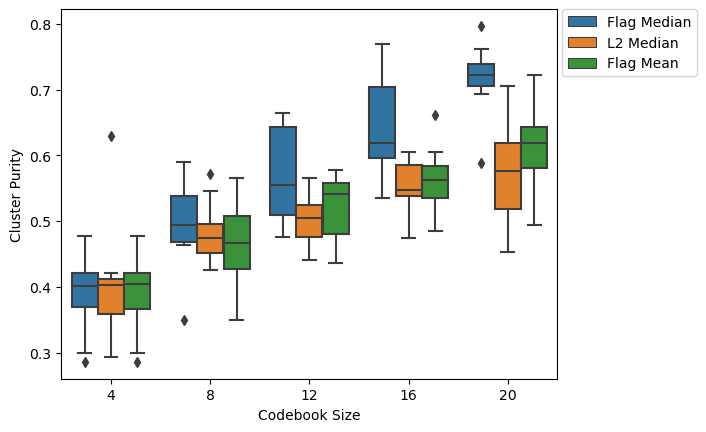}
    \caption[]{An LBG implementation on the Mind's Eye dataset. The results of 3 different implementations of LBG for codebook sizes $4, 8, 12, 16 $ and $20$. The flag median is competitive with the $\ell_2$-median and flag mean for a size $4$ codebook and outperforms $\ell_2$-median and flag mean for codebook sizes $8, 12, 16, 20$.}
    \label{fig:LBG}
\end{figure}

For this example, we use subspaces (points in $\Gr(48,1024)$) that represent clips with action labels bend, follow, pickup, ride-bike and run. There are 27 examples of bend, 32 of follow, 27 of pickup, 17 of ride-bike and 24 of run. We run the Linde-Buzo-Grey (LBG) algorithm \cite{linde1980algorithm,stiverson2019subspace} to cluster these data with different sized codebooks (numbers of centers) and prototype calculation using the flag median, $\ell_2$-median and flag mean. In the LBG algorithm, we calculate distance using chordal distance. For each number of centers, we run 10 trials with different LBG initializations. The results are in Figure \ref{fig:LBG}.

We note that the flag median produces the highest cluster purities for 8, 12, 16 and 20 clusters. In all of the previous experiments we found that the flag median is more robust to outliers which may be the key factor in the success of the flag median prototype LBG implementation. We also note that the $\ell_2$-median and the flag mean LBG implementations have similar cluster purities for each of the codebook sizes. Again, this is consistent with the similar behavior the $\ell_2$-median and the flag mean MNIST experiments (see Figures  \ref{fig:MNIST_classifier} and \ref{fig:1MDS}).

\subsection{UCF YouTube dataset}
Our final dataset is a subset UCF YouTube Action dataset \cite{liu2009recognizing}. This dataset contains 11 categories of actions. For each category, the videos are grouped into groups with common features.  For this expeeriment,  we take approximately one example  from each group within an action category. Specifically our dataset consists of 23 examples of basketball shooting, 22 of biking/cycling, 25 of diving, 24 of golf swinging, 24 of horse back riding, 24 of soccer juggling, 23 of swinging, 24 of tennis swinging, 24 of trampoline jumping, 22 of volleyball spiking, and 24 of walking with a dog. Since these RGB videos are quite large, we convert them to greyscale. Then we generate a matrix for each video whose columns are vectorizations of each frame. Finally, we perform the QR decomposition of each video and take the first $10$ columns of $\Q$ to be it's representative on the Grassmannian.

\begin{figure}[ht]
    \centering
    \includegraphics[width=.8\textwidth]{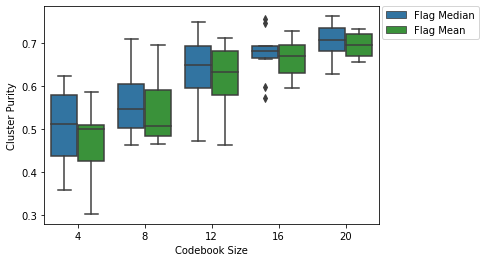}
    \caption[]{An LBG implementation on the YouTube dataset. The results of 2 different implementations of LBG for codebook sizes $4, 8, 12, 16 $ and $20$. The flag median outperforms flag mean for all codebook sizes.}
    \label{fig:LBG_youtube}
\end{figure}

We then run subspace LBG with $48$ dimensional flag mean and the flag median. The results are in Figure \ref{fig:LBG_youtube}. We choose to omit the $\ell_2$-median LBG implementation since the Weiszfeld-type algorithm since it can only compute a $10$ dimensional prototype. We run our LBG implementations with $10$ trials for each of the following codebook sizes: $4,8,12,16$ and $20$. We see the flag median LBG implementation out preform the flag mean LBG implementation in all trials.

\section{Conclusion}
In this paper we presented a new prototype, the flag median, for clusters of points on the Grassmannian. We propose the FlagIRLS algorithm to approximate solutions to the flag median optimization problem. We run experiments comparing the flag median, flag mean, and the $\ell_2$-median. In our experiments, we find the FlagIRLS generally converges faster than gradient descent. In addition, we discover that the flag median is the most robust to outliers and produces higher cluster purities than the flag mean and $\ell_2$-median algorithms.

Future work with the flag median and FlagIRLS could involve machine learning or add details to the mathematical theory. For machine learning, the flag median can be used as a step in a subspace $k$-means algorithm, Grassmannian $n$-shot learning or any other machine learning algorithm in which calculating an ``average'' is a step. Most likely these types of algorithms will be useful for classifying images and videos. In terms of mathematics, we would like to find domain on which the flag median problem is convex and proofs for the convergence rates of FlagIRLS is an open problem. There are potential connections between this flavor of optimization problem and frame theory; so further investigation in this direction could prove useful. Finally, the flag median could be generalized to other spaces such as Stiefel manifolds.
\newline
\\
\textbf{Acknowledgement:} This work was partially supported by National Science Foundation award NSF-ATD 1830676.

\bibliographystyle{unsrt}  
\bibliography{references}

\end{document}